\documentclass[10pt,twocolumn,letterpaper]{article}

\usepackage{cvpr}              

\usepackage{graphicx}
\usepackage{amsmath}
\usepackage{amssymb}
\usepackage{booktabs}
\usepackage{multirow}
\usepackage{dblfloatfix}
\usepackage{float}
\usepackage{graphicx}
\usepackage{subcaption}
\usepackage{verbatim}
\usepackage[export]{adjustbox}

\usepackage[pagebackref,breaklinks,colorlinks]{hyperref}

\usepackage[capitalize]{cleveref}
\crefname{section}{Sec.}{Secs.}
\Crefname{section}{Section}{Sections}
\Crefname{table}{Table}{Tables}
\crefname{table}{Tab.}{Tabs.}

\def\cvprPaperID{12} 
\def\confName{CVPR}
\def\confYear{2023}

\begin{document}

\title{All Keypoints You Need: Detecting Arbitrary Keypoints on the Body of Triple, High, and Long Jump Athletes}

\author{Katja Ludwig \hspace{1.cm} Julian Lorenz \hspace{1.cm} Robin Schön \hspace{1.cm} Rainer Lienhart\\
Chair for Machine Learning and Computer Vision, University of Augsburg\\
{\tt\small \{katja.ludwig, julian.lorenz, robin.schoen, rainer.lienhart\}@uni-a.de}
}
\maketitle

\begin{abstract}
Performance analyses based on videos are commonly used by coaches of athletes in various sports disciplines. In individual sports, these analyses mainly comprise the body posture. This paper focuses on the disciplines of triple, high, and long jump, which require fine-grained locations of the athlete's body. Typical human pose estimation datasets provide only a very limited set of keypoints, which is not sufficient in this case. Therefore, we propose a method to detect arbitrary keypoints on the whole body of the athlete by leveraging the limited set of annotated keypoints and auto-generated segmentation masks of body parts. Evaluations show that our model is capable of detecting keypoints on the head, torso, hands, feet, arms, and legs, including also bent elbows and knees. We analyze and compare different techniques to encode desired keypoints as the model's input and their embedding for the Transformer backbone.
\end{abstract}

\section{Introduction}

Analyzing athletes during trainings or competitions is popular among all sports disciplines. In team sports, typical applications are tactics analyses or the tracking and evaluation of athlete's running paths and the ball during a game. Coaches and athletes of individual sports use video analyses to precisely track the body position. They use these methods to assess the performance of the athletes, to derive improvements, and develop appropriate training routines. The key to analyze the body position is mostly to identify the locations of specific keypoints on the body of an athlete. With these keypoints, exact movements and derived parameters can be evaluated. In this paper, we focus on the disciplines of triple, high, and long jump. Athletes of these disciplines are highly interested in their step frequency, speed, and body posture during the in-run and the jump phase.
\begin{figure}[t]
  \begin{subfigure}{0.49\linewidth}
    \includegraphics[height=4.5cm]{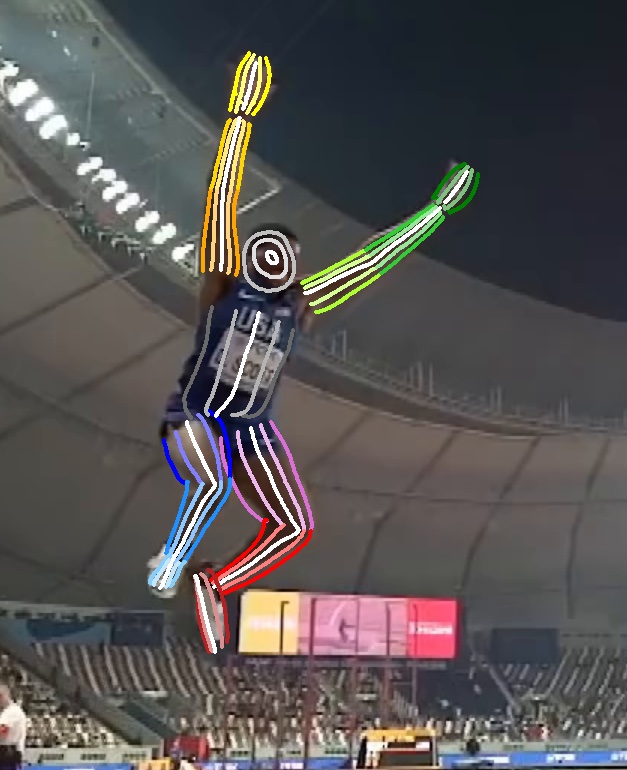}
  \end{subfigure}
  \hfill
  \begin{subfigure}{0.49\linewidth}
    \includegraphics[height=4.5cm, right]{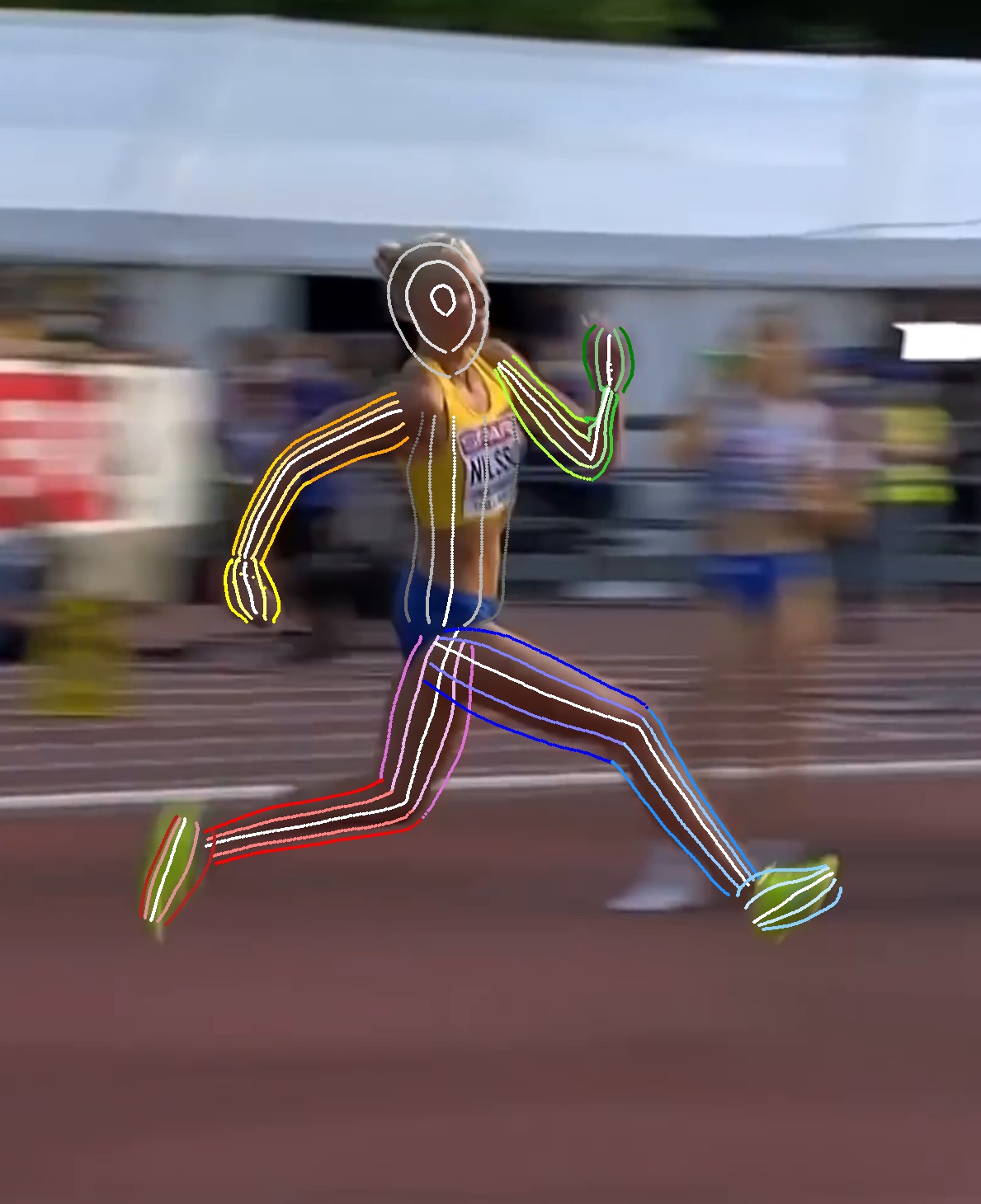}
  \end{subfigure}
  \hfill
\vspace{-0.1cm}    
   \caption{Two detection results of arbitrary keypoints on images of our \emph{jump-broadcast} dataset using our model, visualized with three equally spaced lines to both sides of each body part including the outer boundary in pure color and the central line in white with a color gradient from the boundary to the central line.}
   \label{fig:example_prediction}
   \vspace{-0.4cm}
\end{figure}

Typically, only professional athletes have access to such evaluations as they require a large effort. 2D Human Pose Estimation (HPE) models can help to improve the availability of such video analyses, because they can automate the keypoint detection part which is commonly the most time consuming task. However, the fixed set of keypoints that standard 2D HPE models provide is not sufficient for some kinds of analyses. Adding more keypoints to a fixed set requires annotating all new keypoints in most images, which is again very time consuming and the resulting model is still limited to a (larger) fixed set of keypoints. In order to remove such restrictions, recent works \cite{ludwig2023detecting, ludwig2022recognition} have proposed a technique to detect arbitrary keypoints on the limbs of humans. They adapt the Vision Transformer (ViT) \cite{visiontransformer} based method TokenPose \cite{tokenpose} by replacing the learned tokens for each fixed keypoint with a query embedding. This query embedding encodes the keypoint that should be detected. Query embeddings are generated from human readable query inputs. After various Transformer layers, the output of the model corresponding to these query embeddings is converted to small heatmaps from which the final keypoint detection is obtained. However, these methods are currently limited to the limbs, which we try to overcome in this paper. Figure \ref{fig:example_prediction} shows that our model is capable of detecting arbitrary points on the head, torso, arms, and legs including hands and feet. It also correctly estimates keypoints on bent limbs like elbows and knees.
The contributions of this work can be summarized as follows:

\begin{itemize}
\item We enlarge the area of the body for which arbitrary keypoints can be detected by the feet, hands, torso, and head. We further improve this model such that it can detect keypoints on bent elbows and knees correctly.

\item We propose different techniques for the model to encode the desired target keypoints. We encode the head either based on a reference line or based on an angle. Additionally, we use an encoding based on coordinates of a normalized pose. We further apply different embedding methods to the encoded input.

\item We release the \emph{jump-broadcast} dataset, a new dataset with 2403 annotated triple, high, and long jump athletes sampled from 27 hours of 26 different TV broadcast videos. All images are annotated with 20 keypoints. The dataset further contains 1797 automatically generated segmentations masks. The dataset is available here: \url{https://www.uni-augsburg.de/en/fakultaet/fai/informatik/prof/mmc/research/datensatze/}

\item Experiments on the \emph{jump-broadcast} dataset and a second triple and long jump dataset prove that our methods are capable of detecting any desired keypoint on the body of athletes. We improve the evaluation scheme of previous work by using more keypoints and provide a detailed evaluation of the performance of keypoints located on the different body parts. Our best approach for the jump-broadcast dataset is available here: \url{https://github.com/kaulquappe23/all-keypoints-jump-broadcast} 
\end{itemize}

\section{Related Work}

2D HPE  is a popular technique for computer vision aided analyses in individual sports. Recently, Transformer \cite{transformer} based 2D HPE architectures are gaining popularity, while CNN architectures like the High Resolution Net (HRNet) \cite{hrnet} are still very common. TokenPose \cite{tokenpose} is a Transformer architecture that achieves the best results with the first three stages of an HRNet as feature extractor, but there exists also a pure Transformer variant. All variants make use of the ViT \cite{visiontransformer} architecture, which cuts an image or a feature map into small patches which serve as the input sequence to the Transformer. ViTPose \cite{vitpose} proves that pure ViT based HPE models are also capable of achieving SOTA scores by adding a decoder with deconvolutions after the ViT layers. The HRFormer \cite{hrformer} architecture combines the ideas of the HRNet and the ViT and uses ViT layers while maintaining branches of different feature resolutions like the HRNet. 
Leveraging the idea of focusing on the part of the image where the person is located, Zeng et al. \cite{tokenclusteringtransformer} cluster tokens of less important image areas like the background, while keeping many tokens for important areas. Ma et al. \cite{tokenpruning} follow a similar idea by deleting the tokens of unimportant image areas. They achieve a more lightweight architecture as a consequence.
Apart from that, focusing on multi-person pose estimation, Shi et al. \cite{end2endhpe} propose a fully end-to-end framework based on Transformers. 

Video analyses during training or competitions are common for most professional sports athletes. Analyses in team sports often involve tracking and identification of the ball (and alike) and the players on the field in order to analyze their trajectories, e.g. in (ice) hockey \cite{icehockey, sportstracking}, soccer \cite{soccernet}, volleyball, football, or basketball \cite{sportstracking}. The tracking task is challenging in the domain of sports, as the athletes are similarly dressed, move fast and are often occluded or out of view of the camera. In individual sports, typical analyses involve estimating keypoints on the human body and sports equipment. Liu et al. \cite{badminton} detect the badminton shuttle and reconstruct its 3D trajectory from monocular videos. Table tennis stroke types and the poses of the athletes are further detected by Kulkarni et al. \cite{tabletennis}. Stepec et al. \cite{skijumpstyle} use estimated poses of ski jumpers and their trajectories in order to automatically generate the style score. Hudovernik et al. \cite{jiujitsu} estimate the poses of competing Jiu-Jitsu athletes to automatically detect combat positions and derive the scores of the athletes. Since the footwork is really important in fencing, Zhu et al. \cite{fencing} estimate the poses of fencers and classify fine-grained footwork actions of the athletes.

\section{Jump-Broadcast Dataset}

\newcommand{\mysize}{3.2cm}
\begin{figure*}[htb]
  \begin{subfigure}{0.15\linewidth}
	\centering    
    \includegraphics[height=\mysize]{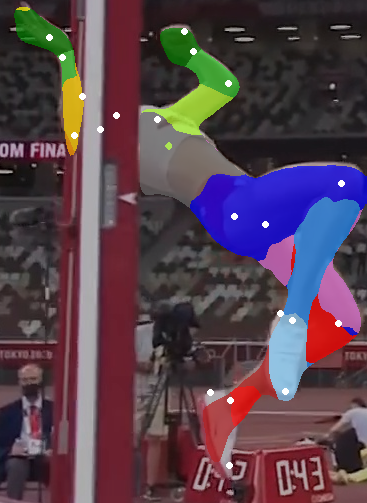}
  \end{subfigure}
  \hfill
  \begin{subfigure}{0.11\linewidth}
  \centering
    \includegraphics[height=\mysize]{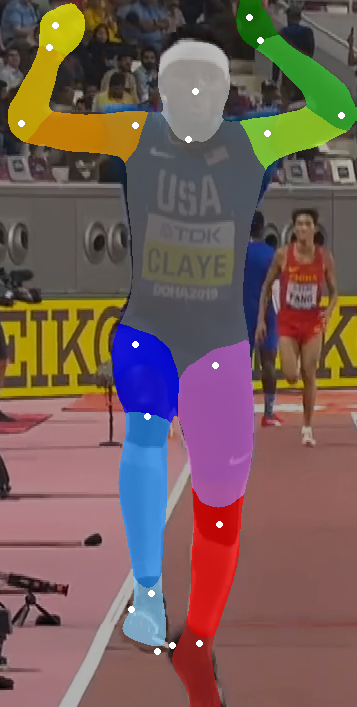}
  \end{subfigure}
  \hfill
  \begin{subfigure}{0.11\linewidth}
  \centering
    \includegraphics[height=\mysize]{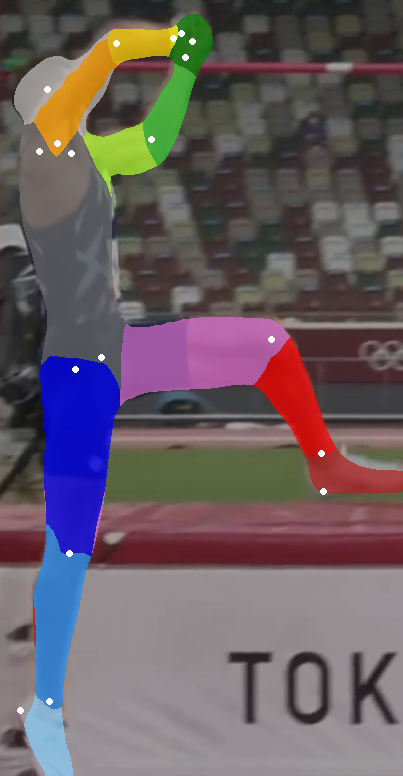}
  \end{subfigure}
  \hfill
  \begin{subfigure}{0.25\linewidth}
  \centering
    \includegraphics[height=\mysize]{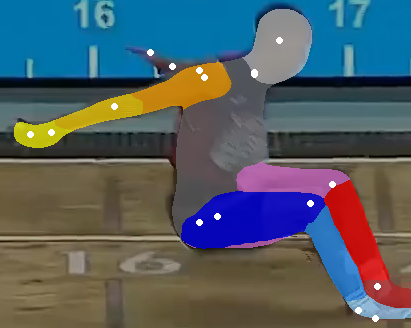}
  \end{subfigure}
  \hfill
  \begin{subfigure}{0.16\linewidth}
  \centering
    \includegraphics[height=\mysize]{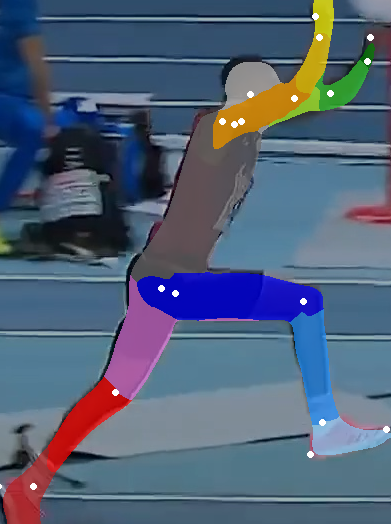}
  \end{subfigure}
  \hfill
  \begin{subfigure}{0.16\linewidth}
  \centering
    \includegraphics[height=\mysize]{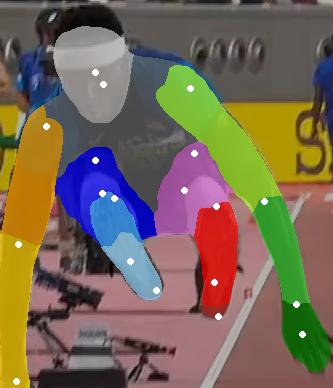}
  \end{subfigure}
  \hfill
  \caption{Cropped example images from the \emph{jump-broadcast} dataset. Segmentation masks are displayed as an overlay. Annotated keypoints are visualized in white. These examples show the variety of poses in our dataset, including occlusions, front and side views, the in-run, and extreme poses during the jump phase.}
  \label{fig:jump-broadcast_dataset}
  \vspace{-0.3cm}
\end{figure*}
We release the \emph{jump-broadcast} dataset to enable a public benchmark on arbitrary keypoint detection for triple, high, and long jump athletes. We have collected 26 videos of competitions from broadcast TV footage, summing up to 27 hours of video material. 9 videos cover triple jump competitions, 8 videos long jump competitions and the remaining 9 videos high jump competitions. A total of 193 different male and female athletes are present in the video footage. The sports sites, lighting conditions and image quality vary throughout our dataset. Moreover, it contains a lot of extreme poses, especially during the jump phase. We select the frames by sampling approx. 5 equidistant frames from each jump (including in-run and jump phase) and each camera perspective. Slow motion replays are mostly recorded with a different camera and therefore seen as a new camera perspective. We select 2403 images in total and annotate them with the following 20 keypoints: head, neck, left/right shoulder, left/right elbow, left/right wrist, left/right hand, left/right hip, left/right knee, left/right ankle, left/right heel, left/right toe tip. Furthermore, the annotations include information whether a frame corresponds to a slow motion replay and the name of the athlete as is presented in the TV broadcast. We split the dataset in 1805 images for training, 576 images for testing and 122 images for validation in such a way that each athlete is only included in a single subset, even if they have participated in multiple competitions. 

We use the DensePose \cite{densepose} framework from detectron2 \cite{detectron2} to automatically generate segmentation masks for our dataset. Since some images are very blurry and the athletes perform extreme poses in comparison to everyday activities, some masks are completely or partly wrong. We sort out the worst segmentation masks by hand but keep masks that are partly correct. The advantage of our approach is that it can deal with partly correct segmentation masks in most cases. In the end, we keep 1797 segmentation masks, 1338 belonging to the training set, 97 to the validation set and 362 to the test set containing the head, torso, left/right upper arm, left/right forearm, left/right hand, left/right thigh, left/right lower leg and left/right foot. Figure \ref{fig:jump-broadcast_dataset} visualizes some images with generated segmentation masks.

\section{Method}

All variants are based on the TokenPose \cite{tokenpose} architecture. In a first step, this architecture takes an image and feeds it through the first three stages of an HRNet for feature extraction. These feature maps are then split into visual feature patches and embedded to create visual tokens via a linear projection. The visual tokens are fed jointly with keypoint query tokens through multiple Transformer layers. In the end, the output of the ViT corresponding to the keypoint query tokens is transformed to heatmaps via a shared MLP. The final keypoint coordinates are then retrieved from the heatmaps with the DARK \cite{dark} method. In contrast to the pure TokenPose architecture, we do not learn representations for the keypoint query tokens, but use transformations to embed human readable encodings in the token domain. Our method can be applied to any TokenPose variant.

\subsection{Ground Truth Generation}

For the limbs, we follow the strategy described in \cite{ludwig2022recognition, ludwig2023detecting} to generate ground truth keypoints. At first, we draw a vector from one enclosing keypoint to the other, which we will call $v_e$. Second, an orthogonal line $l_o$ to $v_e$ is created and the furthest points $c_l$ and $c_r$ that lie on $l_o$ and the segmentation mask are retrieved. $c_l$ is always located on the left side relative to the orientation of $v_e$ and $c_r$ on the right side. Arbitrary keypoints are then located on the line between the so-called intersection points $c_l$ and $c_r$. We further add the constraint that $c_l$ and $c_r$ need to be part of a coherent area of the segmentation mask in order to deal with small errors in our automatically generated segmentation masks, like the left thigh in the fourth image of Figure \ref{fig:jump-broadcast_dataset}. 
We incorporate the same technique for the feet, using the toe tip and the heel as the keypoints that enclose the body part. For the torso, we use the neck as the first enclosing keypoint and the virtual keypoint in the middle of the hip keypoints as the second one. Regarding the hands, this technique has the drawback that it only detects keypoints that lie between the enclosing keypoints. If we would use the hand and wrist keypoint as enclosing keypoints, our model would not be able to detect the finger tips. Since the finger tips are not annotated, we extend the line through the wrist and the hand keypoint beyond the hand keypoint to the boundary of the hand segmentation mask and use this point instead of the hand keypoint as the second enclosing keypoint during ground truth generation. See Section \ref{sec:head} for more details.


\subsubsection{Head}\label{sec:head}

\begin{figure}[t]
  \begin{subfigure}{0.48\linewidth}
\centering    
    \includegraphics[height=2.5cm]{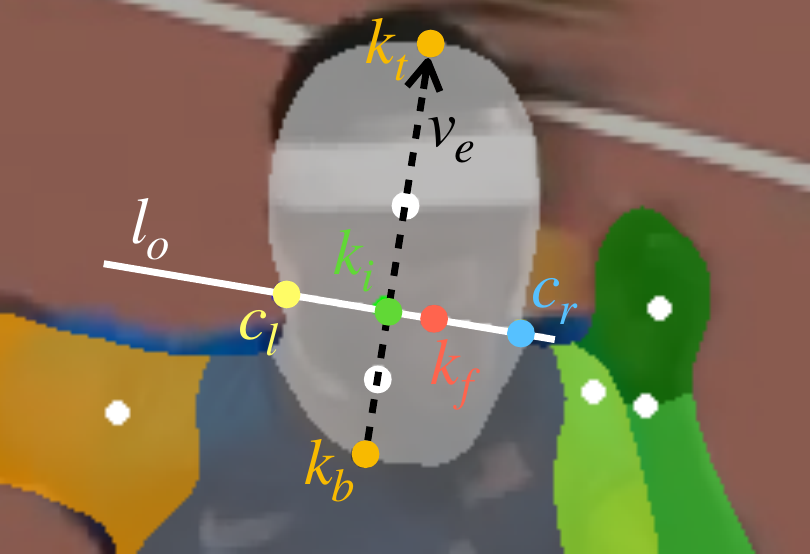}
  \end{subfigure}
  \hfill
  \begin{subfigure}{0.48\linewidth}
\centering    
    \includegraphics[height=2.5cm]{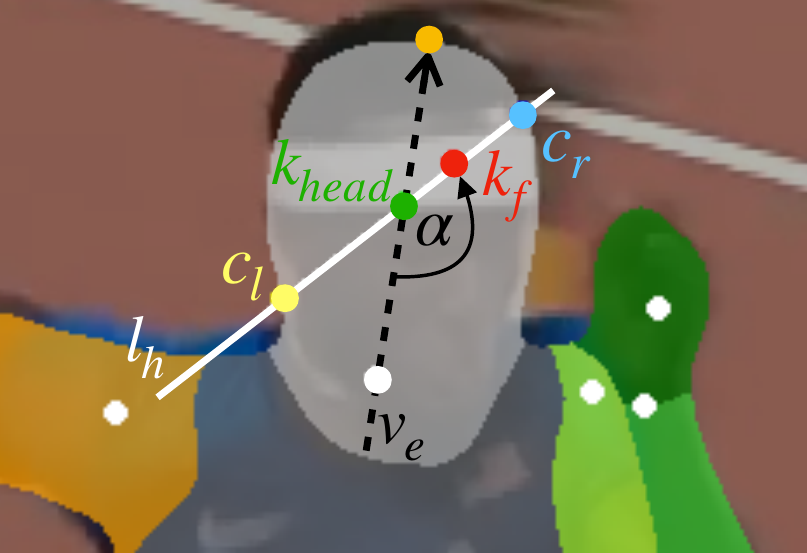}
  \end{subfigure}
  \hfill
   \caption{Two versions of generating ground truth keypoints on the head. On the left, the line through the head and neck keypoint is extended to the head boundary (orange), an orthogonal line to this line is drawn (white) and a keypoint on this line between the boundary points of the segmentation mask (blue and yellow) is chosen (red). On the right, a line (white) rotated around the head keypoint (green) at a random angle is generated and a keypoint on this line between the boundary points of the segmentation mask (blue and yellow) is selected (red).}
   \label{fig:head}
   \vspace{-0.3cm}
\end{figure}

Standard keypoints that are usable as enclosing keypoints for the head are neck and the head keypoint itself. In contrast to other datasets, where the head keypoint is located at the  top of the head, the head keypoint is located in the middle of the head in our datasets. Therefore, we cannot use it as an enclosing keypoint, in this case we would only be able to detect keypoints on the lower half of the head. Hence, we use the strategy already described for the hands, which is also visualized in Figure \ref{fig:head} on the left. At first, the line $v_e$ through the neck and head keypoint is created and extended to the head top $k_t$ and the chest $k_b$, visualized with orange dots in Figure \ref{fig:head}. We select a random intermediate point $k_i$ on $v_e$ (visualized with a green dot in Figure \ref{fig:head}), draw the orthogonal line $l_o$ to this line, determine the intersection points $c_l$ and $c_r$ with the boundary of the segmentation mask and select a random point $k_f$ between $k_i$ and $c_{l/r}$. We call this the \emph{extension} method.

Since the head is rather round, the creation process close to the head top seems counterintuitive. The distance between $c_l$ and $c_r$ is decreasing and even approaching $0$, the closer $k_i$ is to $k_t$. Hence, we propose a second strategy to generate keypoints on the head. We use the head keypoint as the center, choose a random angle $\alpha \in [0, 2\pi)$ and rotate a line $l_h$ counterclockwise around the center according to $\alpha$, while $\alpha = 0$ corresponds to $v_e$. Then, we continue as before with detecting $c_l$ and $c_r$ as the intersection of $l_h$ with the boundary of the segmentation mask and randomly choosing a point $k_f$ between the head keypoint and $c_{l/r}$, which is visualized in Figure \ref{fig:head} on the right. $c_r$ is used for angles $\alpha \in [0, \pi)$ and $c_l$ for $\alpha \in [\pi, 2\pi)$. Since we rotate for a certain angle, we refer to this as the \emph{angle} method.

\subsubsection{Bent Limbs: Elbows and Knees}

\begin{figure}[b]
\vspace{-0.2cm}  
  \centering
  \begin{subfigure}{0.6\linewidth}
\centering    
    \includegraphics[width=\linewidth]{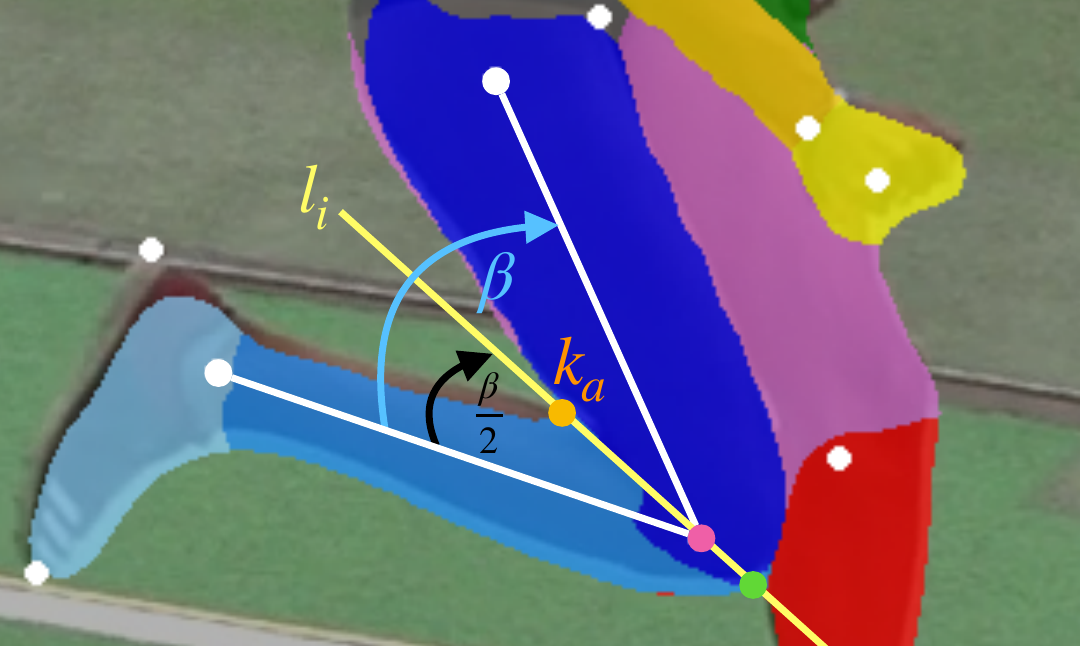}
  \end{subfigure}
  \hfill
   \caption{Visualization of the anchor generation at a bent knee. We generate a line with half the bending angle $\beta$ through the knee keypoint, visualized in yellow. Then, we determine the intersection of that line with the segmentation mask boundary (orange and green) and select the anchor point within the acute angle, so the anchor is the orange point in this image.}
   \label{fig:anchor}
\end{figure}
With the keypoint generation strategy on the limbs as used in \cite{ludwig2022recognition, ludwig2023detecting}, arbitrary keypoints on the upper arm, forearm, thigh, and lower leg could be detected if they lie in between the enclosing keypoints. This works well in most cases, but not if the limbs are strongly bent, which happens frequently in the case of triple, high, and long jump. Therefore, we adapt the generation strategy to fit also the case of bent elbows and knees. For that purpose, we determine the point on the inner side of the bent joint at first, which we call the \emph{anchor point} or $k_a$ in the following. To retrieve that point, we calculate the bending angle $\beta$ of the joint, which is the angle enclosed by the line through the hip and the knee and the line through the knee and the ankle or the angle enclosed by the line through the shoulder and the elbow and the line through the elbow and the wrist. Then, we generate a line $l_i$ through the knee or elbow keypoint with half of the bending angle. We determine the intersections of that line with the boundary of the segmentation mask and set the anchor $k_a$ as the point on the side with the acute angle. We unify the segmentation mask for the lower and upper body part in this case.
See Figure \ref{fig:anchor} for details.
\begin{figure}[t]
  \centering
  \begin{subfigure}{0.6\linewidth}
\centering    
    \includegraphics[width=\linewidth]{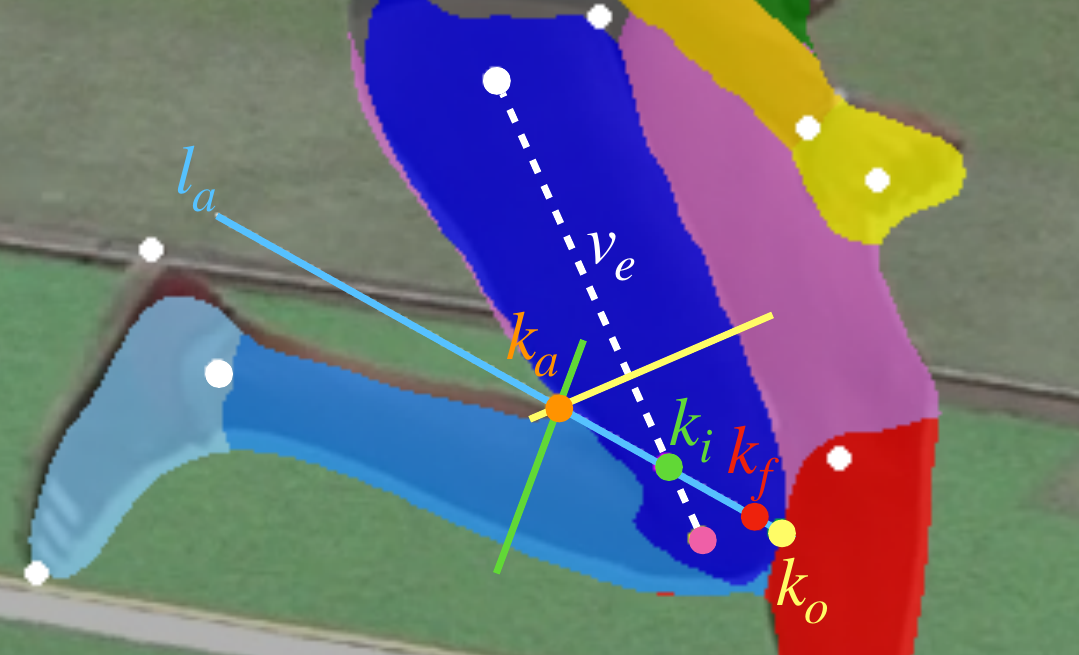}
  \end{subfigure}
  \hfill
   \caption{Visualization of the keypoint generation on a bent knee. The anchor point is visualized in orange, the line $l_a$ in blue. The intersection point $k_i$ is colored green and the point $k_o$ yellow. The final keypoint is visualized in red. The line $l_a$ always lies somewhere between the yellow and the green line.}
   \label{fig:knee}
   \vspace{-0.2cm}
\end{figure}
In the next step, visualized in Figure \ref{fig:knee}, we rotate a line $l_a$ for a random angle around the anchor point, starting in a $90^\circ$ angle to the upper body part (yellow line in Figure \ref{fig:knee}) and stopping at a $90^\circ$ angle to the lower body part (green line in Figure \ref{fig:knee}). We determine the intersection point $k_i$ of $l_a$ with the vector $v_e$ through the enclosing keypoints of the corresponding body part (upper or lower one) and chose a random final keypoint $k_f$ either between $k_i$ and $k_a$ or between $k_i$ and $k_o$. We do not discriminate between the segmentation mask for upper and lower body part in the whole process since the boundary is often detected rough and it is somehow not clearly defined where the thigh/upper arm ends and the lower leg/forearm begins. With this strategy, arbitrary points on bent limbs can be generated.

\subsection{Query Encoding}

With the described strategies, it is possible to generate arbitrary ground truth points. However, it is impossible to maintain an infinite set of keypoints in the way that a heatmap is generated for every possible keypoint, like described in \cite{ludwig2022recognition, ludwig2023detecting}. Therefore, we use a different strategy and tell the model which keypoints it should detect with a special query encoding which is part of the model's input. Since ViT architectures can deal with inputs of various length, it is possible to query for an arbitrary number of keypoints. In this paper, we evaluate two fundamentally different query encodings and two nuances for one variant.

\subsubsection{Vectors}

The fist approach uses multiple vectors to encode the desired keypoint. Let the dataset contain $n$ keypoints, then the first vector called \emph{keypoint vector} has length $n$ and the second vector called \emph{thickness vector} has length $3$. For the limbs and the feet, we follow the strategy described in\cite{ludwig2022recognition} to fill the vectors. Let $k_1, k_2$ be two keypoints enclosing a body part and $k_i$ be the orthogonal projection of the desired keypoint on the line $l_o $ between $k_1$ and $k_2$. Then $k_i = p\cdot k_1 + (1-p) \cdot k_2$. We set the entries belonging to $k_1$ and $k_2$ in the keypoint vector to $p$ and $1-p$, respectively. For the torso, we set the keypoint vector entry belonging to the neck to $p$ and the entries belonging to the left and right hip joint to $0.5(1-p)$, since we use the middle of the two hip keypoints as the reference point. In case of the hands and the extension method for the head, where we extend the line through the reference keypoints $k_1$ and $k_2$ to the points $k_t$ and $k_b$, $k_i = p\cdot k_t + (1-p)\cdot k_b$, but we set the entries in the keypoint vector in the same way as before. The thickness vector is created similarly, let the final keypoint be $k_f = q\cdot c_{1/2} + (1-q) \cdot k_i$, then we set the middle entry of the thickness vector to $1-q$ and the first or last entry to $q$, depending if $c_l$ or $c_r$ is chosen \cite{ludwig2022recognition}. 

If the angle method is used for the head, this approach is not applicable as the creation logic is completely different. Therefore, we introduce a third vector called the \emph{angle vector}. It has just one entry which is set to $0$ in case of all other body parts, meaning that it is not used. For the head, its value indicates the rotation angle $\alpha$ in percent, starting with the line through head and neck keypoint and rotating counterclockwise around the head keypoint like described in Section \ref{sec:head}. This means that the lines $l_h$ for rotation angle percentages $0 < p <= 0.5$ and $0.5 + p$ are identical. They differ in their corresponding intersection point. $c_r$ is used for percentage $p$ and $c_l$ for $0.5 + p$. The final keypoint can now be calculated as $k_f = q\cdot c_{l/r} + (1-q)\cdot k_{head}$, and we set the thickness vector to $[0, 1-q, q]$ no matter which intersection point is used as this information is already encoded in the angle vector. We further set the entry of the keypoint vector corresponding to the head keypoint to $1$. 

For elbows and knees, we adapt the encoding slightly. In these cases, $k_i$ is not projected perpendicular (along $l_o$) on the line between $k_1$ and $k_2$, but along the line $l_a$ and we further use $l_a$ to determine the intersection points $c_l$ and $c_r$ and the thickness percentage $q$.

\subsubsection{Normalized Pose}

\begin{figure}[t]
  \centering
  \begin{subfigure}{0.6\linewidth}
\centering    
    \includegraphics[width=\linewidth]{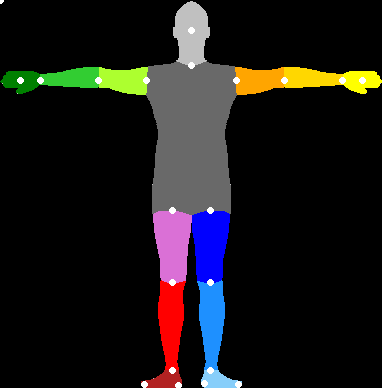}
  \end{subfigure}
  \hfill
   \caption{Visualization of the normalized pose. All used body parts are colored and the fixed keypoints are visualized in white. }
   \label{fig:normpose}
\end{figure}

The second approach involves a normalized human pose, similar to \cite{ludwig2022recognition}. The normalized human pose contains the keypoints and the body part segmentation masks of a human in a T-shaped pose, visualized in Figure \ref{fig:normpose}. We use feet turned outwards and hands downwards such that the maximum area of these body parts is visible in the segmentation masks. Each desired keypoint is now represented with the normalized $x$- and $y$-coordinate of this pose. Normalized coordinates mean that the upper left corner has coordinates $(0, 0)$ and the lower right corner $(1, 1)$. Hence, each keypoint is characterized by only two values in this encoding approach.

\subsection{Keypoint Token Embedding}

Each keypoint encoding, no matter which of the three described options is used, needs to be converted to a token of the same size as the visual tokens in order to be compatible with the ViT. This process is called \emph{embedding}. The straightforward way is to use a linear layer to convert a vector of arbitrary length to a vector with the desired length. In case of the vector approach, we have more than one vector in the encoding. Hence, we have the option to concatenate the vectors to one vector at first and then embed them, or to embed them at first to tokens with a half or a third of the embedding size and concatenate them afterwards. Furthermore, apart from using a single linear layer, we try to enhance the embeddings by adding more layers and ReLU operations.

\section{Experiments}

The base architecture for all our experiments is TokenPose-Base \cite{tokenpose} with an input image size of $192 \times 256$, an embedding dimension of $192$ and a conversion of the final embeddings to heatmaps of size $48 \times 64$. Athletes are cropped before the images are fed through the network. We pretrain our model on the COCO dataset \cite{coco} and finetune it on our individual sports related datasets. Random flipping, random rotation of up to $45^\circ$, random scaling in the range of $[0.65, 1.35]$, and color jitter are applied as augmentation techniques during training.

\subsection{Evaluation Metrics}

We use the Percentage of Correct Keypoints (PCK) and the Percentage of Correct Thickness (PCT) as evaluation metrics. Both metrics calculate the percentage of all desired keypoints that the network has correctly predicted at a threshold $t$. The PCK uses the distance between the left shoulder and right hip keypoints times the threshold as the maximum distance for a correct prediction. We use $t=0.1$ and $t=0.05$ in our evaluations, which corresponds to approx. 6cm and 3cm, respectively. The PCT calculates the differences between the predicted thickness and the ground truth thickness for keypoints on all body parts analogous to the keypoint generation by subtraction or addition depending if the ground truth and predicted keypoint are located on the same or different sides of the intermediate point/the head keypoint. Correct predictions are the ones with thickness differences below $t$. We use $t=0.2$ in our evaluations.

\cite{ludwig2022recognition} uses a random number of up to 50 freely chosen keypoints per image during evaluation. \cite{ludwig2023detecting} increases their evaluation strategy to a fixed number of 200 randomly created keypoints. In this paper, we aim to evaluate the full range of possible keypoints together with the different encoding and embedding strategies for certain body parts. Hence, on each body part, we create a specific set of evaluation keypoints: we use thickness $0, 0.5$ and $1$ and create 25 equally spaced keypoints with these thicknesses to both sides of the body part, resulting in 125 keypoints per body part. During evaluation, we treat elbows and knees as individual body parts, too. Hence, if all body parts are completely visible, this results in a total number of 2250 keypoints per image. The benefit of this strategy is that it is reproducible and hence provides a comparable metric for future work. It also weights all body parts equally and allows a separate comparison of the results for each body part. 

\subsection{Collapsing Elbow and Knee Keypoints}

Elbows and knees connect the upper arm with the forearm and thigh with the lower leg, including them in our method enables retrieving continuous keypoints along the whole arms and legs.
Especially in these cases, it is very important to interpret $v_e$ as a vector and define $c_l$ and $c_r$ relative to its direction.
\begin{figure}[tb]
  \centering
  \begin{subfigure}{0.49\linewidth}
    \includegraphics[height=5.3cm, left]{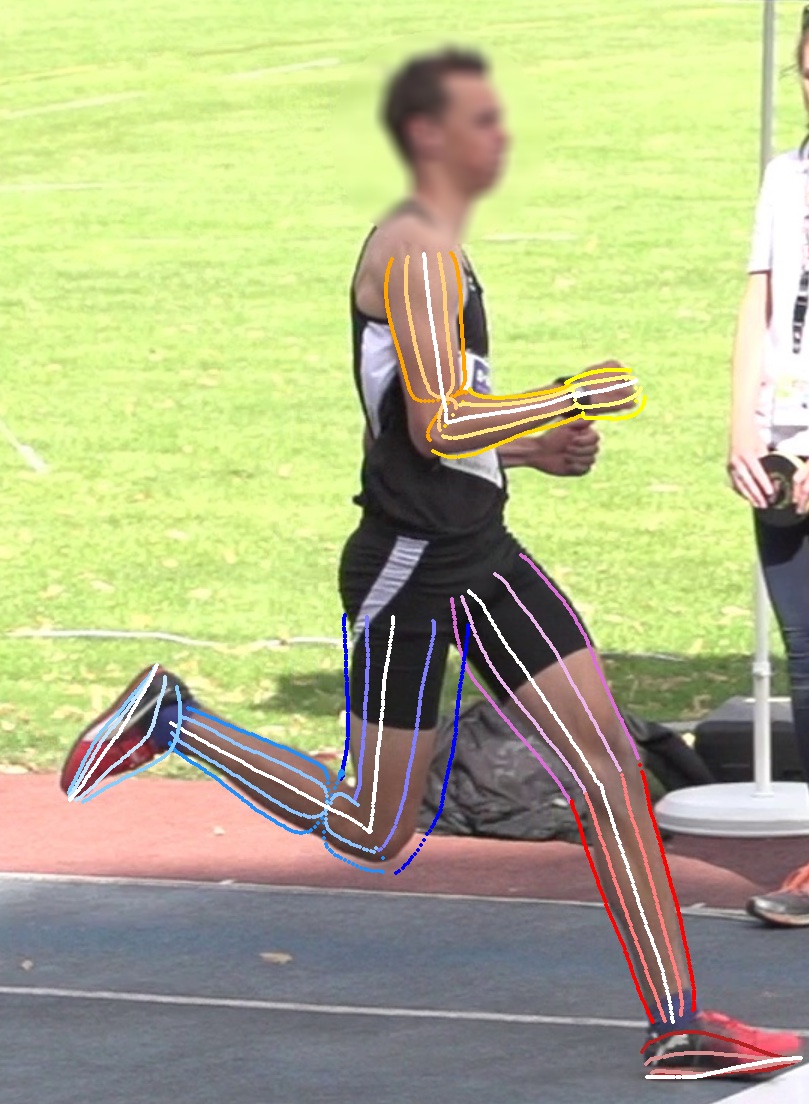}
  \end{subfigure}
  \hfill
  \begin{subfigure}{0.49\linewidth}
    \includegraphics[height=5.3cm, right]{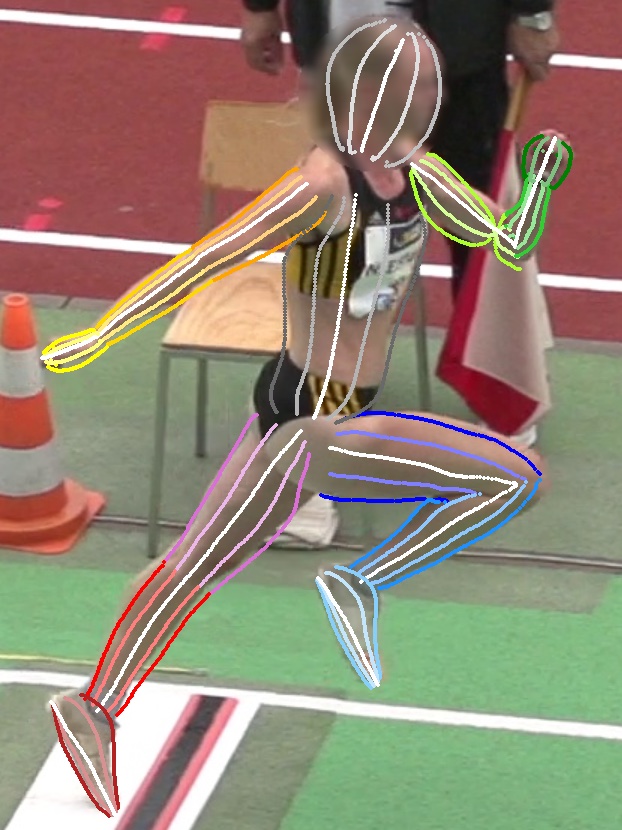}
  \end{subfigure}
  \hfill
   \caption{Collapsing points around the right elbow and knee (first image) and the left elbow (second image) in the case that $c_l$ and $c_r$ swap sides between the lower and upper body part.}
   \label{fig:leftright_collapse}
   \vspace{-0.2cm}
\end{figure}
This ensures that $c_l$ and $c_r$ are always located on the same side of the upper body part (thigh or upper arm) and the lower body part (lower leg or forearm). If $c_l$ and $c_r$ swap the side between the lower and upper body part, we can encounter a collapse of the keypoints around the swapping location, visualized in Figure \ref{fig:leftright_collapse}.

\subsection{Triple and Long Jump Dataset}

\begin{table*}[tb]
\begin{center}
\resizebox{\linewidth}{!}{ 
\begin{tabular}{c|cc|ccc|cccccccccc}
\toprule
Enc. & Emb. & Con. & PCK@0.1 & PCK@0.05 & PCT & head & torso & u.arm & elbow & f.arm & hand & thigh & knee & l.leg & foot\\
\midrule
V-E & 1 & 1 & 94.1 & 82.1 & 69.9  & 81.0  & 76.0  & 77.7  & 66.1  & 67.9  & 55.1  & 82.2  & 71.7  & 77.7  & 49.9  \\
V-E & 2 & 1 & 94.2 & 82.3 & 71.0  & 82.0  & 76.6  & 78.7  & 68.2  & 68.6  & 56.7  & 82.7  & 71.9  & 77.9  & 52.4  \\
V-E & 2 & 0 & 94.0 & 82.0 & 70.3  & 81.6  & 74.9  & 78.1  & 67.2  & 68.4  & 55.4  & 82.2  & 71.8  & 78.3  & 51.3  \\
V-E & 3 & 1 & 94.2 & 82.4 & 71.0  & 81.6  & 76.6  & 78.4  & 68.0  & 68.9  & 57.1  & 82.7  & 71.9  & 78.6  & 52.2  \\
\midrule
V-A & 1 & 1 & 94.3 & 82.6 & 71.6  & 94.4  & 75.9  & 78.9  & 68.3  & 68.5  & 55.7  & 82.7  & 71.9  & 78.0  & 51.2  \\
V-A & 2 & 1 & 94.3 & 82.5 & 71.6  & 92.8  & 76.1  & 78.8  & 68.0  & 69.1  & 56.3  & 82.3  & 71.9  & 78.2  & 51.0  \\
V-A & 2 & 0 & 94.5 & 82.7 & 71.9  & 93.2  & 76.9  & 79.0  & 68.4  & 69.1  & 56.2  & 82.8  & 71.8  & 78.3  & 52.3  \\
V-A & 2 & 2 & 94.3 & 82.2 & 71.5  & 92.5  & 76.5  & 78.5  & 68.0  & 68.8  & 56.4  & 82.4  & 72.0  & 77.8  & 51.6  \\
V-A & 3 & 1 & 94.3 & 82.4 & 71.5  & 94.4  & 76.0  & 78.4  & 67.9  & 68.4  & 56.1  & 82.8  & 71.4  & 78.0  & 51.2  \\
\midrule
NP & 1 & - & 92.5 & 79.7 & 66.6  & 94.6  & 72.2  & 73.6  & 63.2  & 63.4  & 48.9  & 81.7  & 70.8  & 73.7  & 33.6  \\
NP & 2 & - & 92.6 & 78.9 & 65.5  & 94.2  & 71.2  & 73.2  & 62.8  & 63.0  & 48.8  & 81.1  & 70.0  & 72.3  & 27.6  \\
NP & 4 & - & 92.5 & 79.6 & 66.8  & 94.5  & 73.0  & 73.4  & 63.3  & 62.5  & 48.0  & 81.7  & 71.3  & 75.4  & 34.8  \\
 \bottomrule
\end{tabular}
}
\end{center}
\vspace{-0.2cm}
\caption{Recall values for the triple and long jump test set in \% at PCK@$0.1$ and PCK@$0.05$ and PCT@$0.2$. These scores are evaluated on the test set with the fixed keypoints and on the described 2250 keypoints per body part as far as these keypoints exist in the image. The first column indicates the used encoding: \emph{V} indicates the vector approach, either with the head extension strategy (symbolized with \emph{V-E}) or the head angle strategy (\emph{V-A}). \emph{NP} refers to the normalized pose approach. The second column names the number of layers used for the embedding and the third column the number of layers that is executed before concatenation of the vectors. The third table section contains the average metric results over all keypoints in the test set, the fourth section lists the PCT scores separately for each keypoint type, left and right keypoints are combined. } \label{tab:results1}
\end{table*}
\renewcommand{\mysize}{3.7cm}
\begin{figure*}[b]
  \begin{subfigure}{0.17\linewidth}
	\centering    
    \includegraphics[height=\mysize]{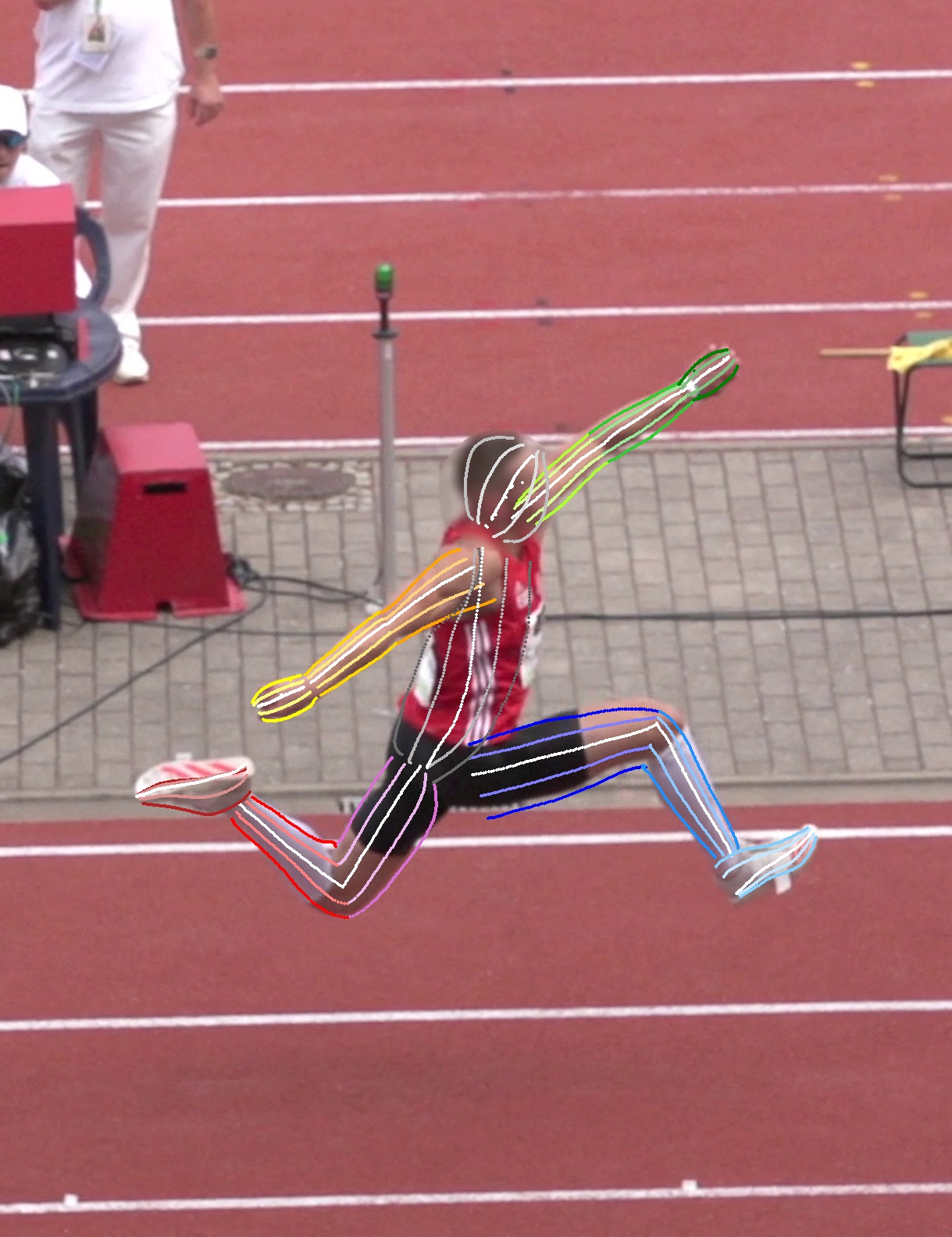}
  \end{subfigure}
  \hfill
  \begin{subfigure}{0.16\linewidth}
  \centering
    \includegraphics[height=\mysize]{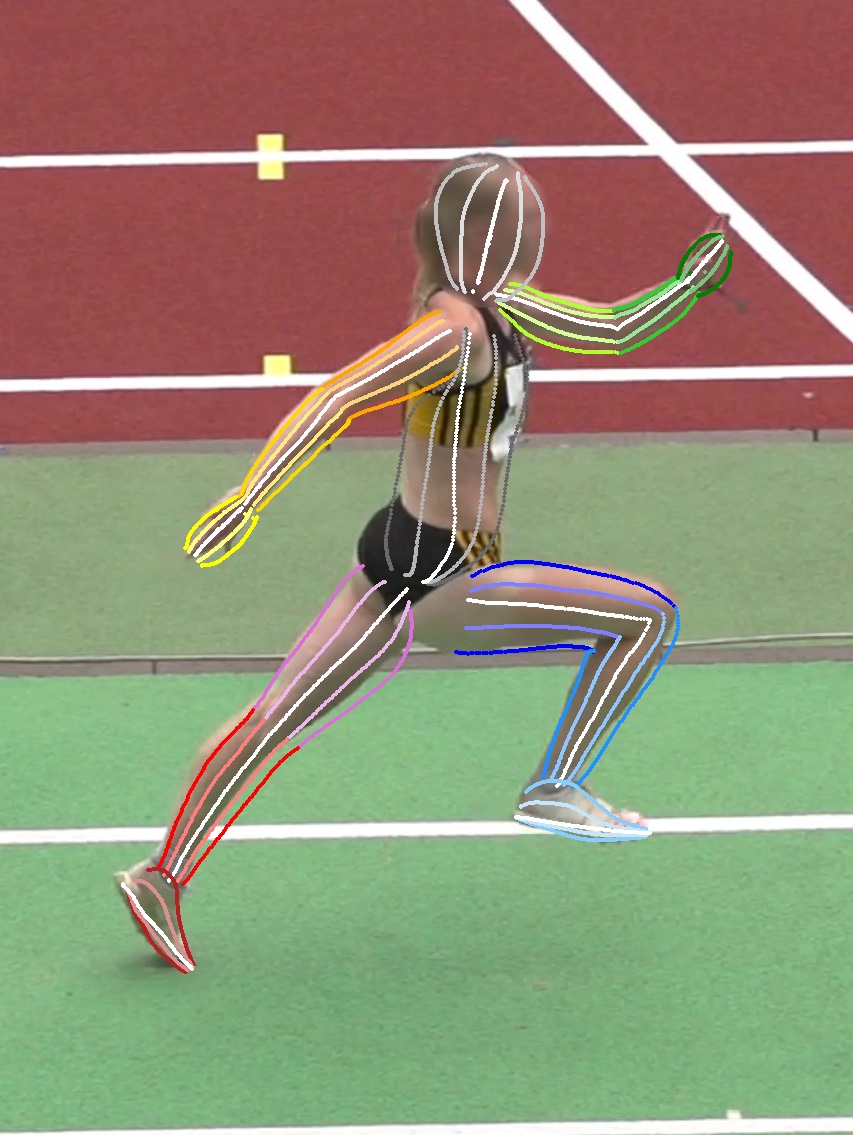}
  \end{subfigure}
  \hfill
  \begin{subfigure}{0.16\linewidth}
  \centering
    \includegraphics[height=\mysize]{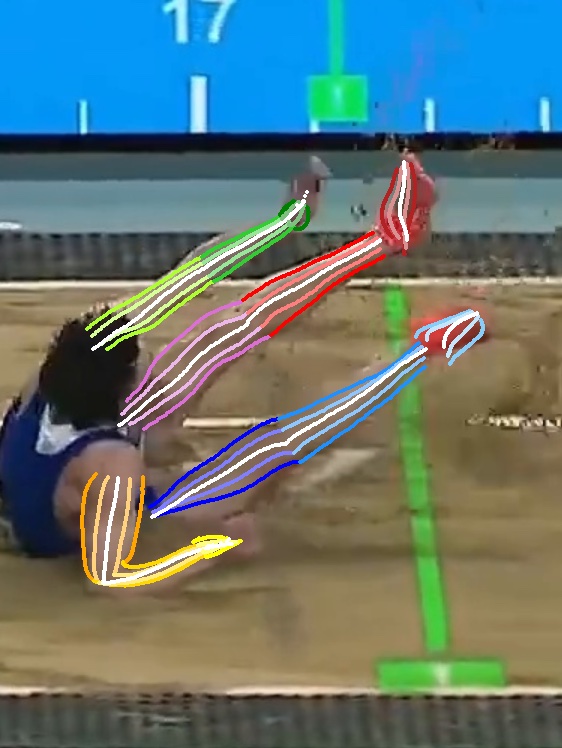}
  \end{subfigure}
  \hfill
  \begin{subfigure}{0.16\linewidth}
  \centering
    \includegraphics[height=\mysize]{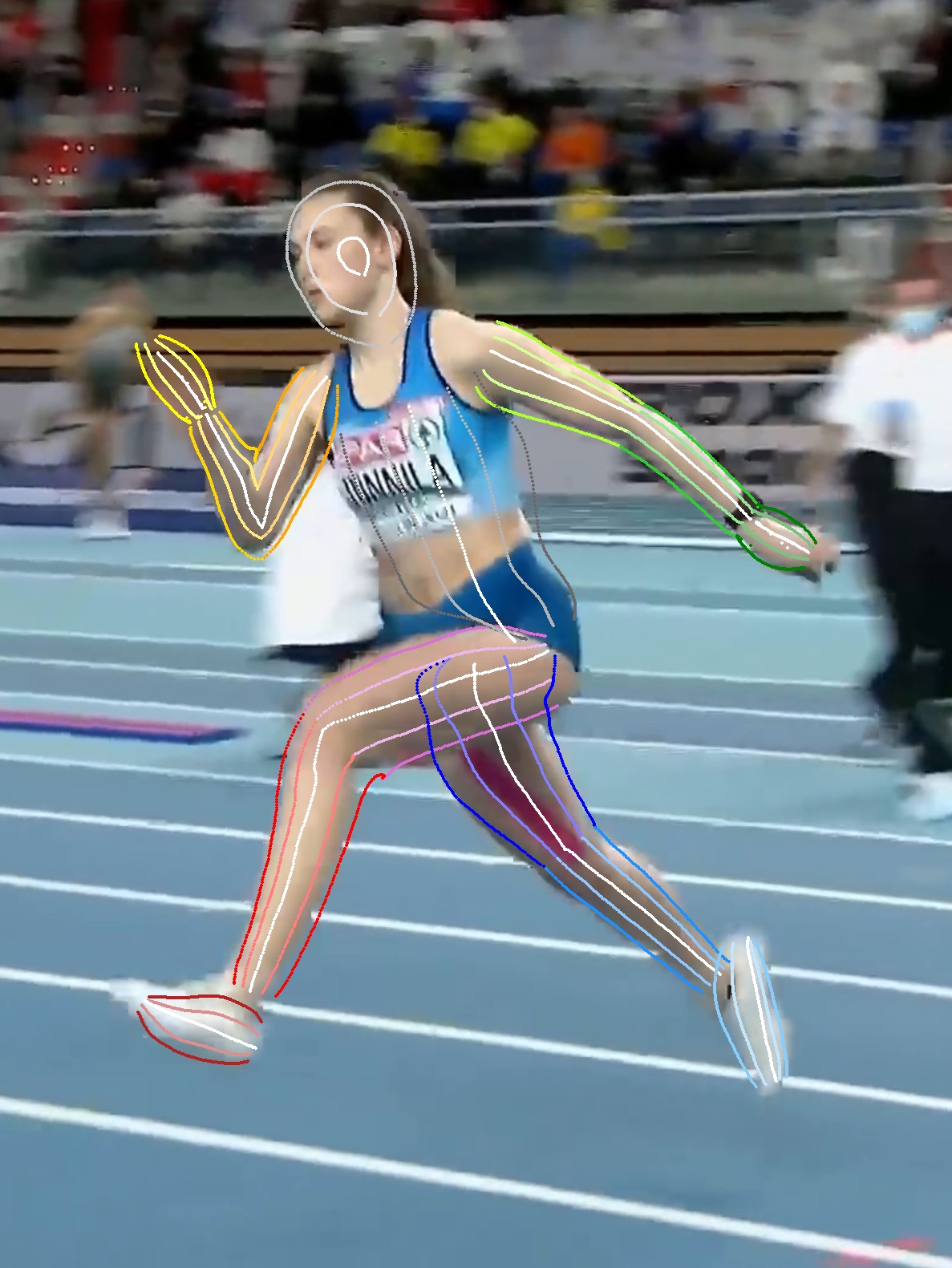}
  \end{subfigure}
  \hfill
  \begin{subfigure}{0.16\linewidth}
  \centering
    \includegraphics[height=\mysize]{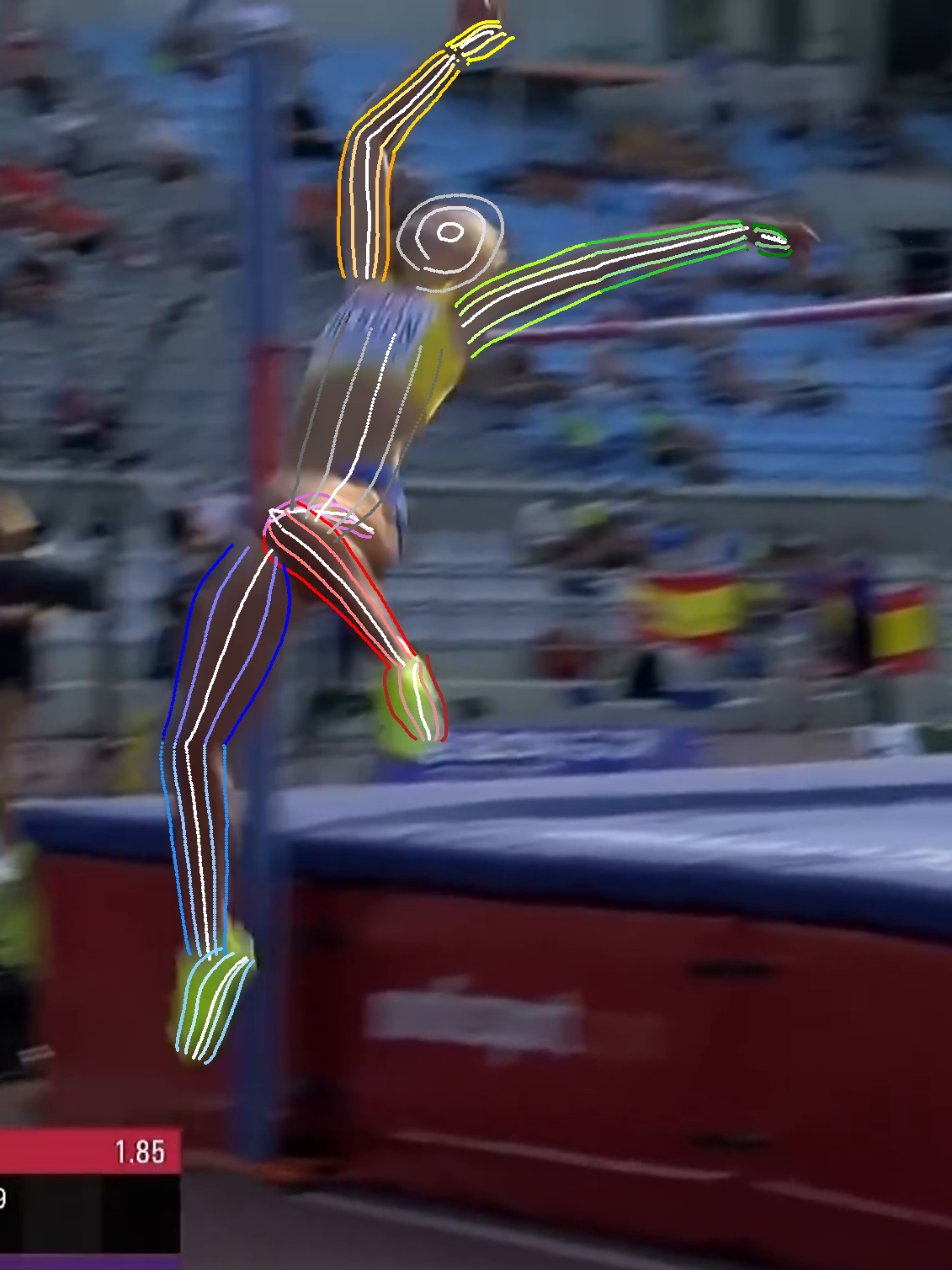}
  \end{subfigure}
  \hfill
  \begin{subfigure}{0.16\linewidth}
  \centering
    \includegraphics[height=\mysize]{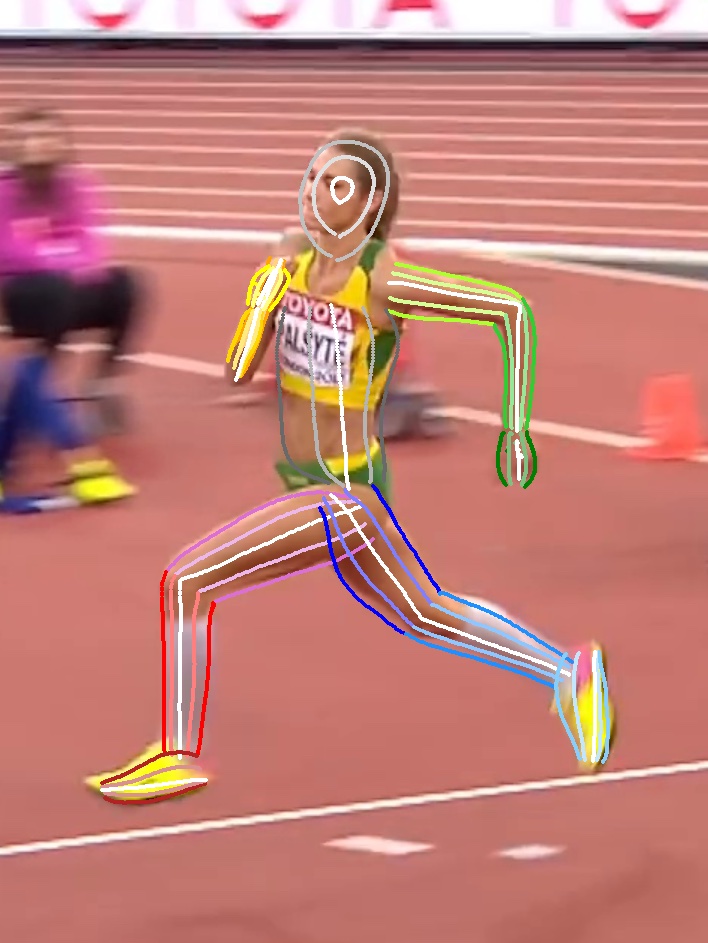}
  \end{subfigure}
  \hfill
  \caption{Detection results for images of the triple and long jump dataset (first two images) and the \emph{jump-broadcast} dataset (last four images), visualized with equally spaced lines to both sides of each body part including the outer boundary in pure color and the central line in white with a color gradient from the boundary to the central line. The first two images are generated with the vector encoding using the extension strategy and the last four images with the angle strategy for the head in order to show the differences between the strategies. Occluded/overlapping body parts are omitted for clarity.}
  \label{fig:results}
\end{figure*}
We first evaluate our methods on the triple and long jump dataset. The quality of its images is higher than that of the jump-broadcast dataset since the cameras were installed specifically for the analysis tasks and not for TV broadcasts. It contains 4101 training images, 1462 testing images and 464 validation images from various sports sites, indoor and outdoor locations, and from a variety of different athletes during triple and long jump competitions or trainings. It is labeled with the same 20 keypoints as the jump-broadcast dataset. Segmentation masks are also obtained with detectron2 \cite{detectron2} and kept for all images.

We execute four experiments with different embedding strategies using the vector encoding with the extension technique for the head, five experiments with the angle technique and three experiments with the normalized pose encoding. The results are displayed in Table \ref{tab:results1}. Overall, the different strategies achieve similar results. All are capable of detecting arbitrary keypoints on all body parts. Generally, the approaches with the vector encoding achieve better results in most body parts. 
The number of layers in the embedding process does also not make a large difference independent from the encoding type, which is in contrast to the results reported in \cite{ludwig2022recognition}. The position of the concatenation also does not show a difference. For the angle strategy, earlier concatenation performs slightly better, for the endpoint strategy, later concatenation is better. Hence, these differences are not significant and might just be due to typical training performance variance. A significant difference is observed regarding the head keypoint. The angle strategy achieves an absolute improvement of over 10\% for all experiments. Although the normalized pose approach achieves worse results for all other body parts, its performance on the head keypoint is as good as the vector endcoding with the angle strategy.
The detection scores vary largely among the different body parts. The feet and the hands seem the most challenging body parts, since the PCT is at most 52.4\% and 57.1\%, respectively. Head, thighs, upper arms and lower legs are the body parts with the best scores over all experiments. These results might be explained by the fact that these body parts generally lead to larger (and therefore also thicker) segmentation masks, which make it easier for the network to learn and detect precise keypoints. Furthermore, their boundaries are easier to determine than for (maybe bent) elbows and knees or hands and feet. 
The normalized pose approach has especially difficulties with the hands and feet which might be due to the fact that the area of these body parts has the lowest size in the normalized pose template.
The best overall PCT and PCK@0.05 score is achieved with the vector encoding and the angle approach and a two layer embedding with concatenation right at the beginning.
Qualitative results for this dataset are visualized in the first two images in Figure \ref{fig:results}.

\subsection{Jump-Broadcast}

\begin{table*}[tb]
\begin{center}
\resizebox{\linewidth}{!}{ 
\begin{tabular}{c|cc|ccc|cccccccccc}
\toprule
Enc. & Emb. & Con. & PCK@0.1 & PCK@0.05 & PCT & head & torso & u.arm & elbow & f.arm & hand & thigh & knee & l.leg & foot\\
\midrule
V-E & 1 & 1 & 90.5 & 71.8 & 61.3  & 69.8  & 79.6  & 65.8  & 48.1  & 51.4  & 38.5  & 79.5  & 64.5  & 67.5  & 42.8  \\
V-E & 2 & 1 & 90.8 & 72.1 & 62.5  & 71.3  & 80.2  & 67.6  & 49.4  & 53.5  & 40.2  & 80.4  & 64.9  & 66.2  & 47.9  \\
\midrule
V-A & 1 & 1 & 91.0 & 71.9 & 63.1  & 83.5  & 80.4  & 66.4  & 49.1  & 53.0  & 39.2  & 80.2  & 64.4  & 66.8  & 46.0  \\
V-A & 2 & 1 & 91.2 & 72.1 & 63.3  & 83.0  & 79.3  & 67.7  & 49.2  & 52.6  & 40.4  & 80.8  & 65.0  & 66.1  & 46.8  \\
V-A & 2 & 2 & 91.0 & 71.9 & 63.0  & 82.7  & 79.8  & 66.6  & 48.0  & 53.9  & 40.6  & 80.2  & 63.4  & 66.1  & 47.1  \\
\midrule
NP & 1 & - & 88.0 & 67.9 & 57.9  & 83.5  & 77.3  & 54.6  & 41.1  & 44.8  & 32.6  & 79.8  & 63.3  & 65.8  & 31.7  \\
NP & 4 & - & 88.1 & 67.9 & 58.1  & 83.4  & 77.9  & 56.5  & 41.1  & 45.8  & 33.3  & 79.0  & 63.0  & 65.0  & 31.5  \\
 \bottomrule
\end{tabular}
}
\end{center}
\caption{Recall values for the triple and long jump test set in \% at PCK@$0.1$ and PCK@$0.05$ and PCT@$0.2$. These scores are evaluated on the test set with the fixed keypoints and on the described 2250 keypoints per body part as far as these keypoints exist in the image. The first column indicates the used encoding: \emph{V} indicates the vector approach, either with the head extension strategy (symbolized with \emph{V-E}) or the head angle strategy (\emph{V-A}). \emph{NP} refers to the normalized pose approach. The second column names the number of layers used for the embedding and the third column the number of layers that is executed before concatenation of the vectors. The third table section contains the average metric results over all keypoints in the test set, the fourth section lists the PCT scores separately for each keypoint type, left and right keypoints are combined.} \label{tab:results2}
\end{table*}
For the jump-broadcast dataset, we execute two experiments with the vector encoding and the extension strategy, three experiments with the angle strategy and two experiments with the normalized pose encoding. The results are shown in Table \ref{tab:results2}. Since the quality of the images is worse than for the triple and long jump dataset, the scores are generally lower. 
The effect that the results of the angle strategy improve the score for the head can be observed for this dataset as well. The improvement is even slightly larger, with an absolute increase of over 11\% for all variants. The score for the other body parts are similar across all variants of the extension encoding and the angle encoding. Using more layers or other concatenation techniques shows no significant difference. Keypoints on the hands and feet stay the most challenging ones in this dataset. For the vector encoding approaches, the scores for the feet are better than the scores for the hands, which is different from the results of the triple and long jump dataset. However, this changes for the normalized pose approach, which might be due to the fact that the feet are slightly smaller in the pose template than the hands. Moreover, the scores of the elbow are significantly lower relative to the score of the upper arm and forearm compared to the other dataset. The best scores are achieved for the head, the thigh, and the torso. 
The normalized pose approach achieves again the lowest overall scores. The largest drop is also observable for keypoints on the feet, while the score for keypoints on the head are better than for the extension strategy. Furthermore, the angle strategy achieves the best overall results (PCT, PCK@0.05 and PCK@0.1) with a two layer embedding and the concatenation after one layers. Example detections for this method are displayed in the last four images in Figure \ref{fig:results}.

\section{Conclusion}

In this paper, we propose a method to detect arbitrary keypoints on the whole human body for athletes of triple,  long, and high jump. We introduce a new dataset consisting of images from these disciplines in order to provide a publicly available dataset for future comparisons. This \emph{jump-broadcast} dataset comprises 2403 images from 27 hours of 26 different TV broadcast videos showing 193 athletes. We obtain 1797 segmentation masks of the head, torso, l./r. upper arm, l./r. forearm, l./r.  hand, l./r. thigh, l./r. lower leg, and l./r. foot.

We extend the methods introduced in \cite{ludwig2022recognition, ludwig2023detecting} to make our model capable of estimating arbitrary keypoints on all body parts. To generate ground truth keypoints for the hands, which have no second annotated enclosing keypoint, we extend the line through wrist and hand keypoint to the boundary of the segmentation mask to obtain a second point. For the head, we use either the same technique, or we rotate a line around the head keypoint. We further calculate the points on the inner side of the elbow and knee joints and rotate around these points in order to generate arbitrary points on the elbow and knee, since these joints are often heavily bent during jumps and could not be estimated correctly until now. We evaluate our model with different encodings of the arbitrary keypoints, either as vectors with the extension strategy for the head or the angle strategy, or as keypoint coordinates of a normalized human pose. In the vector case, we introduce a third vector in case of the angle strategy and evaluate the performance depending on the number of embedding layers used before we concatenate the embeddings of the single vectors. Evaluations on the triple and long jump dataset and our newly introduced jump-broadcast dataset show that the results for all variants are generally similar. The normalized pose approach achieves slightly lower scores regarding all experiments. Concatenating the vector embeddings later results in minimal better scores. For both datasets, keypoints on the hand are the most challenging. The extension and angle approach for the head have strengths for different keypoints. For the triple and long jump dataset, the extension approach with a single layer embedding achieves the best overall score while the angle approach with a two-layer embedding leads to the best performance for the jump-broadcast dataset. Hence, the method proposed in this paper is capable of detecting arbitrary keypoints on the whole human body of triple, long, and high jump athletes, including bent elbows and knees.

{\small
\bibliographystyle{ieee_fullname}
\bibliography{references}
}

\end{document}